\documentclass[10pt,twocolumn,letterpaper]{article}

\usepackage[pagenumbers]{cvpr} 
\usepackage[accsupp]{axessibility}

\usepackage{graphicx}
\usepackage{amsmath,amssymb,amsthm,mathabx,amsfonts}
\usepackage{bbm}
\usepackage{url}
\usepackage{algorithmic}
\usepackage{acronym}
\usepackage{enumitem}
\usepackage{balance}
\usepackage{booktabs}
\usepackage[table,xcdraw]{xcolor}
\usepackage{xspace}
\usepackage{setspace}
\usepackage[skip=3pt]{subcaption}
\usepackage[skip=3pt]{caption}
\usepackage[misc]{ifsym}
\usepackage{url}
\usepackage{multirow}
\usepackage{tablefootnote}
\usepackage{colortbl}
\usepackage{xr}

\usepackage[square,sort,comma,numbers]{natbib}
\definecolor{mygray}{gray}{.92}

\makeatletter
\renewcommand{\paragraph}{%
  \@startsection{paragraph}{4}%
  {\z@}{0ex \@plus 0ex \@minus 0ex}{-1em}%
  {\hskip\parindent\normalfont\normalsize\bfseries}%
}
\makeatother

\makeatletter
\newcommand{\thickhline}{%
    \noalign {\ifnum 0=`}\fi \hrule height 1pt
    \futurelet \reserved@a \@xhline
}

\DeclareMathOperator*{\argmax}{arg\,max}

\DeclareMathOperator{\img}{\rm img}

\acrodef{nlp}[NLP]{natural language processing}
\acrodef{plm}[PLM]{pretrained masked language model}
\acrodef{sota}[SOTA]{state-of-the-art}
\acrodef{bs}[BS]{Beam Search}
\acrodef{mhs}[MHS]{Metropolis-Hastings Sampling}
\acrodef{hs}[HS]{Hybrid Search}
\acrodef{uas}[UAS]{unlabeled attachment score}
\acrodef{dda}[DDA]{Directed Dependency Accuracy}
\acrodef{sota}[SOTA]{state-of-the-art}
\acrodef{pos}[POS]{part-of-speech}
\acrodef{asr}[ASR]{attacking success rate}
\acrodef{ppl}[PPL]{Perplexity score}
\acrodef{mcmc}[MCMC]{Markov Chain Monte Carlo}
\acrodef{pcfg}[PCFG]{Probablistic Context Free Grammar}
\acrodef{cpcfg}[C-PCFG]{Compound PCFG}
\acrodef{vcpcfg}[VC-PCFG]{Visually Grounded Compound PCFG}
\acrodef{diora}[DIORA]{Deep Inside-Outside Recursive Autoencoders}
\acrodef{sdiora}[S-DIORA]{Single Tree DIORA}
\acrodef{ner}[NER]{Named Entity Recognition}
\acrodef{cnf}[CNF]{Chomsky Normal Form}
\acrodef{vl}[VL]{vision-language}
\acrodef{sg}[SG]{scene graph}
\acrodef{pt}[PT]{parse tree}
\acrodef{dt}[DT]{dependency tree}
\acrodef{dmv}[DMV]{Dependency Model with Valence}
\acrodef{cl}[CL]{contrastive learning}
\acrodef{vg}[VG]{visual grounding}
\acrodef{vqa}[VQA]{visual question answering}
\acrodef{dda}[DDA]{directed dependency accuracy}
\acrodef{uda}[UDA]{undirected dependency accuracy}
\acrodef{amt}[AMT]{Amazon Mechanical Turk}
\acrodef{mle}[MLE]{Maximum Likelihood Estimation}
\acrodef{hit}[HIT]{human intelligence task}
\acrodef{em}[EM]{expectation-maximization}

\newcommand{\obj}{{\small \texttt{OBJECT}}\xspace}
\newcommand{\attr}{{\small \texttt{ATTRIBUTE}}\xspace}
\newcommand{\rel}{{\small \texttt{RELATIONSHIP}}\xspace}
\definecolor{myyellow}{RGB}{255,255,204}

\usepackage[pagebackref,breaklinks,colorlinks]{hyperref}

\usepackage[capitalize]{cleveref}
\crefname{section}{Sec.}{Secs.}
\Crefname{section}{Section}{Sections}
\Crefname{table}{Table}{Tables}
\crefname{table}{Tab.}{Tabs.}
\crefname{equation}{Eqn.}{Eqns.}

\def\cvprPaperID{2769} 
\def\confName{CVPR}
\def\confYear{2022}

\begin{document}

\title{Unsupervised Vision-Language Parsing: Seamlessly Bridging Visual Scene Graphs with Language Structures via Dependency Relationships}



\author{Chao Lou$^{1,2}$\thanks{Equal contribution. Author orders are coin clipped. This work was conducted when Chao Lou and Yuhuan Lin were research interns at BIGAI. } , Wenjuan Han$^{1}$\thanks{Corresponding author.} , Yuhuan Lin$^3$, Zilong Zheng\textsuperscript{1}\footnotemark[1] \\
\textsuperscript{1} Beijing Institute for General Artificial Intelligence (BIGAI), Beijing, China \\
\textsuperscript{2} ShanghaiTech University, Shanghai, China \\
\textsuperscript{3} Tsinghua Unversity, Beijing, China \\
{\small {\tt louchao@shanghaitech.edu.cn, hanwenjuan@bigai.ai}} \\
{\small {\tt lin-yh20@mails.tsinghua.edu.cn,  zlzheng@bigai.ai}} \\
{\small \url{https://github.com/bigai-research/VLGAE}}
}

\maketitle

\begin{abstract}

Understanding realistic visual scene images together with language descriptions is a fundamental task towards generic visual understanding. Previous works have shown compelling comprehensive results by building hierarchical structures for visual scenes~(e.g., scene graphs) and natural languages~(e.g., dependency trees), individually. 
However, how to construct a joint vision-language~(VL) structure has barely been investigated. 
More challenging but worthwhile, we introduce a new task that targets on inducing such a joint VL structure in an unsupervised manner. Our goal is to bridge the visual scene graphs and linguistic dependency trees seamlessly. Due to the lack of VL structural data, we start by building a new dataset \texttt{VLParse}. Rather than using labor-intensive labeling from scratch, we propose an automatic alignment procedure to produce coarse structures followed by human refinement to produce high-quality ones. Moreover, we benchmark our dataset by proposing a contrastive learning~(CL)-based framework VLGAE, short for Vision-Language Graph Autoencoder.
Our model obtains superior performance on two derived tasks, \ie, language grammar induction and VL phrase grounding. Ablations show the effectiveness of both visual cues and dependency relationships on fine-grained VL structure construction. 


\end{abstract}

\section{Introduction}
\label{sec:intro}

Visual scene understanding has long been considered a primal goal for computer vision. Going beyond the success of high-accurate individual object detection in complicated environments, various attempts have been made for higher-order visual understanding, such as predicting an \textit{explainable}, \textit{structured}, and \textit{semantically-aligned} representation from scene images~\cite{wu2007compositional,johnson2015image,jiang2018configurable}. Such representations not only provide fine-grained visual cues for low-level recognition tasks, but have further demonstrated their applications on numerous high-level visual reasoning tasks, \eg, \ac{vqa}~\cite{tu2014joint,zheng2019reasoning}, image captioning~\cite{chen2020say,yao2010i2t}, and scene synthesis~\cite{jiang2018configurable,johnson2018image}.

\begin{figure}[t!]
    \centering
    \includegraphics[width=\linewidth]{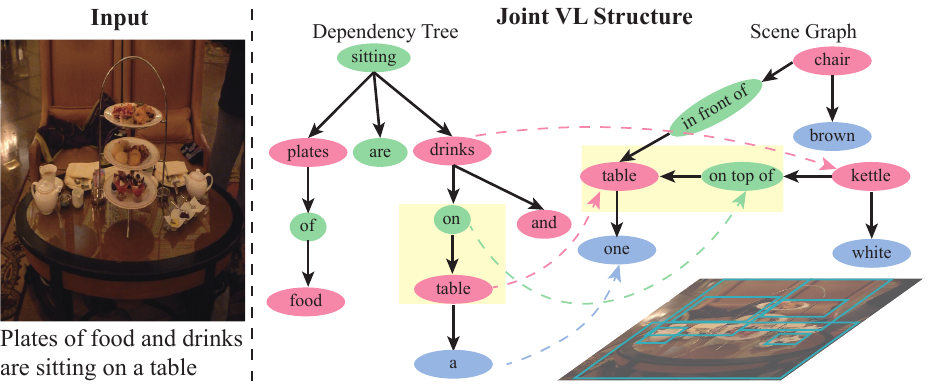}
    \caption{Task illustration of \texttt{VLParse}. Different node types are identified by their background colors and \colorbox{myyellow}{the yellow areas} indicate first-order relationships~($\S$\ref{sec:vl_structure}). 
    }
    \label{fig:task_definition}
    \vspace{-.05in}
\end{figure}
\Ac{sg}, one of the most popular visual structures, serves as an abstraction of objects and their complex relationships within scene images~\cite{johnson2015image,krishnavisualgenome}. Conventional scene graph generation models recognize and predict objects, attributes, relationships, and their corresponding semantic labels purely from natural images in a \textit{fully-supervised} manner~\cite{yang2018graph,suhail2021energy}. Despite the promising performance achieved on large-scale \ac{sg} benchmarks, these methods suffer from limitations on existing datasets and task setting~\cite{gu2019scene}. First, a comprehensive scene graph requires different semantic levels of visual understanding~\cite{li2017scene}, whilst most current datasets only capture a small portion of accessible semantics for classification~\cite{gu2019scene}, which will cause the prediction model bias towards those most-frequent labels. Second, building such datasets requires exhaustive labeling of bounding boxes, relations, and corresponding semantics, which are time-consuming and inefficient. 
Third, it is typically hard to induce a semantically-consistent graphical structure solely from visual inputs, which typically requires an extra visual relation recognition module with heavy manually-labeled supervision. 


Different from dense and noisy visual information, natural language directly provides symbolic and structured information (\eg, grammar) to support the comprehension process. Researches on language structure induction can date back to early computational linguistic theories~\cite{chomsky1956three,chomsky1959certain,chomsky2009syntactic}. Empowered by advances in deep learning techniques, a variety of neural structured prediction algorithms were proposed to analyze more complicated structure information and apply them to natural language tasks~\cite{DIORA,kim2019compound,SDIORA,yang-etal-2020-second,zhang-etal-2021-adapting}. \Ac{dt} parsing, as one essential branch of language structured prediction, aims to generate a parse tree that is composed of vertices representing each word's semantic and syntactic meanings, and directed edges representing the dependency relationships among them. Of note, such tree structure shares a similar idea as in \ac{sg}. However, the ground truth structure (commonly referred to as ``gold structure") requires professional linguists' labeling. To mitigate the data issue, pioneer works have also demonstrated the success of \ac{dt} learning in an unsupervised schema~\cite{kim2019compound,jiang2016unsupervised,jiang-etal-2017-combining}.  

In this work, we leverage the best of both modalities and introduce a new task -- unsupervised \acf{vl} parsing (short for \texttt{VLParse}) -- aiming to devise a joint \ac{vl} structure that bridges visual scene graphs with linguistic dependency trees seamlessly. By ``seamless", we mean that each node in the \ac{vl} structure shall present the well-aligned information of some node in \ac{sg} and \ac{dt}, so are their relationships, as shown in \Cref{fig:task_definition}. To the best of our knowledge, this is the first work that formally defines the joint representation of \ac{vl} structure with dependency relationships. 
Respecting the semantic consistency and independent characteristics, the joint \ac{vl} structure considers both the shared multimodal instances and the independent instances for each modality. In such a heterogeneous graph, semantically consistent instances across two graphs (\ac{dt} and \ac{sg}) are aligned in different levels, which maximizes the retention of the representation from two modalities.
Some previous attempts have shown the benefits of exploring multi-modality information for structured understanding. For example, \citet{shi2019visually} first proposes a visually grounded syntax parser to induce the language structure. \citep{zhao2020visually,zhang2021video} further exploit visual semantics to improve the structure for language. These structures, however, are still for language syntactic parsing rather than for joint vision-language understanding. One closest work to us is VLGrammar~\cite{hong2021vlgrammar}, which builds separate image structures and language structures via compound PCFG~\cite{kim2019compound}. However, the annotations (\ie, segmentation parts) are provided in advance. 

\texttt{VLParse} aims to conduct thoughtful cross-modality understanding and bridge the gap between multiple subtasks: structure induction for the image and language separately and unsupervised visual grounding.
As a complex task, it is comprised of several instances, such as objects, attributes, and different levels of relationships.
The interactions among different instances and subtasks can provide rich information and play a complementary or restrictive role during identification and understanding.

To address this challenging task, we propose a novel \ac{cl}-based architecture, Vision-Language Graph Autoencoder~(VLGAE), aiming at constructing a multimodal structure and aligning \ac{vl} information simultaneously. 
The VLGAE is comprised of feature extraction, structure construction, and cross-modality matching modules. The feature extraction module extracts features from both modalities and builds representations for all instances in \ac{dt} and \ac{sg}. The structure construction module follows the encoder-decoder paradigm, where the encoder obtains a compressed global \ac{vl} representation from image-caption pair using attention mechanisms; the decoder incorporates the inside algorithm to construct the \ac{vl} structure recursively as well as compute the posteriors of spans.
The \ac{vl} structure induction is optimized by \ac{mle} with a negative likelihood loss. 
For cross-modality matching, we compute the vision-language matching score between visual image regions and language contexts. 
We further enhance the matching score with posterior values achieved from the structure construction module. 
This score is used to promote the cross-modality fine-grained correspondence with the supervisory signal of the image-caption pairs via a \ac{cl} strategy; see \Cref{fig:diagram} and \Cref{sec:model} for details.




In summary, our contributions are five-fold: 
(i) We design a joint VL structure that bridges visual \acl{sg} and linguistic \acl{dt}; 
(ii) We introduce a new task \texttt{VLParse} for better cross-modality visual scene understanding~($\S$\ref{sec:task}); 
(iii) We present a two-step \ac{vl} dataset creation paradigm without labor-intensive labelling and deliver a new dataset~($\S$\ref{sec:dataset}); 
(iv) We benchmark our dataset with a novel \ac{cl}-based framework VLGAE~($\S$\ref{sec:model}); 
(v) Empirical results demonstrate significant improvements on single modality structure induction and cross-modality alignment with the proposed framework.

\section{Related Work}
\label{sec:rel_work}
\paragraph{Weakly-supervised Visual Grounding} Visual grounding~(VG) aims to locate the most relevant object or region in an image referred by natural language expressions, such as phrases~\cite{wang2015learning}, sentences~\cite{akula2020words,tu2014joint} or dialogues~\cite{zheng2019reasoning}. 
Weakly-supervised visual phrase grounding, which infers region-phrase correspondences using only image-sentence pairs, has drawn researchers' attention. There are multiple approaches to weakly-supervised visual phrase grounding. \citet{gupta2020contrastive} leverage contrastive learning to train model based on image-sentence pairs data. \citet{wang2020maf} build visually-aware language representations for phrases that could be better aligned with the visual representations. \citet{wang2021improving} develop a method to distill knowledge from Faster R-CNN for weakly supervised phrase grounding. 

Language sentences contain rich semantics and syntactics information. Thus, some researches focus on how to extract and leverage useful information in a sentence to facilitate visual grounding. For example, \citet{xiao2017weakly} use the linguistic structure of natural language descriptions for visual phrase grounding. \citet{yu2018mattnet} learn to parse captions automatically into three modular components related to subject appearance, location, and relationship to other objects, which get rich different types of information from sentences. 
In this work, we propose to induce structures from realistic image-caption pairs without any structure annotations, nor phrase-region correspondence annotations.
Note that different from \citet{wang-etal-2020-maf} who predict the corresponding regions for a given set of noun phrases, noun phrases in \ac{vl} grammar induction are unknown and all spans in the \ac{vl} structure are corresponding regions in the image. 



\paragraph{Language Dependency Parsing} Dependency parsing, a fundamental challenge in \ac{nlp}, aims to find syntactic dependency relations between words in sentences. 
Due to the challenge of achieving gold structures for all available language corpus, unsupervised dependency parsing, whose goal is to obtain a dependency parser without using annotated sentences, has attracted more attention over recent years.
The pioneer work \ac{dmv}~\cite{klein2004corpus} proposes to model dependency parsing as a generative process of dependency grammars.
Empowered by deep learning techniques, NDMV~\cite{jiang2016unsupervised} employ neural networks to capture the similarities between \ac{pos} tags, and learn the grammar based on \ac{dmv}. However, generative models often are limited by independence assumption, so more researchers have paid attention to autoencoder-based approaches~\cite{cai2017crf}, \eg, Discriminative NDMV (D-NDMV)~\cite{han2019enhancing}. 

\paragraph{Visual-Aided Grammar Induction} Visual-aided word representation learning and sentence representation learning achieve positive results.  \citet{shi2019visually} first propose the visually grounded grammar induction task and present a visually-grounded neural syntax learner (VG-NSL). They use an easy-first bottom-up parser~\cite{goldberg-elhadad-2010-efficient} and use REINFORCE~\cite{Williams2004SimpleSG} as gradient estimator for image-caption matching. \citet{zhao-titov-2020-visually} propose an end-to-end training algorithm for Compound PCFG~\cite{kim-etal-2019-compound}, a powerful grammar inducer. \citet{jin-schuler-2020-grounded} formulate a different visual grounding task. They use an autoencoder as a visual model and fuse language and vision features on the hidden states. Different from visually-grounded grammar induction, we not only care about the language structure accuracy but also the fine-grained alignment accuracy.


\section{The \texttt{VLParse} Dataset}\label{sec:dataset}
In this section, we start by formalizing the joint \ac{vl} structure to represent the shared semantics for vision and language. Then we introduce how the dataset, \texttt{VLParse}, is formed in a semi-automatic manner. 

\subsection{Joint Vision-Language Structure}\label{sec:vl_structure}

The \acf{vl} structure is composed of a visual structure \ac{sg}, a linguistic structure \ac{dt} and a hierarchical alignment between \ac{sg} and \ac{dt}.

\paragraph{\Acf{sg}} We define \ac{sg} on an image $\mathbf{I}$ as a structured representation composed of three types of nodes: $\mathcal{T} = \{\obj, \attr, \rel \}$, denoting the image's objects features, conceptual attribute features, and relationship features between two objects. 
Each \obj node is associated with an \attr node; between each pair of \obj nodes, there exists a \rel node.  
Let $\mathcal{R}$ be the set of all relationship types (including ``none" relationship), we can denote the set of all variables in \ac{sg} as $\{v_i^{cls}, v_i^{bbox}, v_i^{type}, v_{i\rightarrow j}; i \neq j \}$, where $v_{i}^{cls}$ is the class label of the $i$-th bounding box, $v_{i}^{bbox} \in \mathbb{R}^4$ denotes the bounding box offsets, $v_{i}^{type} \in \mathcal{T}$ is the node type, and $v_{i\to j} \in \mathcal{R}$ is the relationship from node $v_i$ to $v_j$.

\paragraph{\Acf{dt}} Conventional \ac{dt} is a hierarchy with directed dependency relationships. Given the textual description denoted as a sequence of $N$ words $\mathbf{w} = \{w_1, w_2, ..., w_N\}$, each dependency within \ac{dt} can be denoted as triplet $(w_i, w_j, w_{i\rightarrow j})$, representing a parent node $w_i$, a child node $w_j$ and the direct dependency relationship from $w_i$ to $w_j$, respectively. 
Similar to \ac{sg}, for each node's representation, we additionally append the node's type label $w_{i}^{type} \in \mathcal{T}$. Thus, all variables within in \ac{dt} becomes $\{w_i, w_i^{type}, w_{i \rightarrow j}; i \neq j\}$.

\paragraph{Alignment} The alignment between \ac{dt} and \ac{sg} can be seen as a realization of visual grounding for instances on different levels of the linguistic structure. We hereby define three levels of alignment (see \Cref{fig:task_definition} for illustration):
\begin{itemize}[noitemsep, topsep=0pt]
    \item \textit{Zero-order Alignment}. It defines connections between each node $w_i$ in \ac{dt} with a node $v_i$ in \ac{sg}. 
    
    \item \textit{First-order Alignment}. A first-order relationship can be defined as a triplet $(w_i, w_j, w_{i \rightarrow j})$, including two nodes and a directed dependency. Then the first-order alignment aims to align the triplet in \ac{dt} with a similar triplet $(v_i, v_j, v_{i \rightarrow j})$ in \ac{sg}. 
    
    \item \textit{Second-order Alignment}. A second-order relationship builds upon the first-order relationship and is represented as dependencies among three nodes, \eg, $w_i$, $w_j$, and $w_k$ in \ac{dt}. Similar as the first-order alignment, the second-order alignment aligns such relationships between \ac{dt} and \ac{sg} with similar semantics.
\end{itemize}
\begin{figure}[t!]
    \centering
    \includegraphics[width=\linewidth]{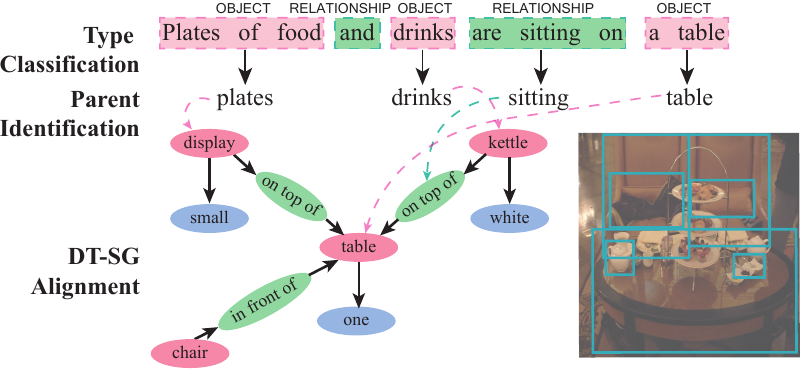}
    \caption{An illustration of the process of the automatic rule-based alignment. After the process of DT rewriting and DT-SG alignment, every instance in DT can be aligned to a SG instance. Through SG, instances in DT can match to image regions.}
    \label{fig:alignment}
    \vspace{-.05in}
\end{figure}

\subsection{Automatic Rule-based Alignment}
In practice, the alignment between \ac{sg} and \ac{dt} is labor-intense and expensive to obtain, whilst unlabeled data is low in cost and large in scale. Thus we design a set of rules to automatically ground the language instances of \ac{dt} to the vision instances of \ac{sg}. This automatic alignment provides beneficial information to reduce the labeling burden on workers. Specifically, we introduce a two-step alignment process, \ie, rule-based \ac{dt} rewriting (DT Rewriting) followed by the alignment between DT and SG (DT-SG Alignment).

\paragraph{DT Rewriting}
We start by introducing the rewriting procedure that extends and deforms \ac{dt} in order to mitigate the difference between \ac{dt} and \ac{sg}. The rewriting considers two modules: 

\noindent\emph{\underline{Type Classification}}\quad{}
We append the type label $x_{i}^{type}$ for conventional DT. More specifically, we label words from DT with three node types as follows:
\begin{itemize}[itemsep=1pt,topsep=2pt]
    \item \obj: The \obj node in \ac{dt} is referred as to a word/phrase that can be grounded to a specific image area. A noun phrase including all words involved except for the attribute is designed as an \obj node. 
    \item \attr: The \attr node is mostly an adjective used to decorate its linked \obj node. In our designed rules, we set words with dependency type \textit{acomp} (adjectival complement) as \attr nodes.
    \item \rel: Two \obj nodes are linked to a \rel node with a directed dependency. 
    \obj nodes are connected to each other through a \rel node. For example in \Cref{fig:alignment}, ``sitting" as a \rel node between two \obj nodes ``drinks" and ``table".
\end{itemize}

\noindent\emph{\underline{Parent Identification}}\quad{} Since it is hard to identify a grounding area in image for an \attr node or a function word (such as the coordinating conjunction and determiner, \etc), we instead define words of these types share the corresponding \obj node's, says parent nodes' grounding area. A parent node is a \textit{noun} representing the core semantics of a noun phrase and \attr nodes, as dependent, modify it. This parent-dependent relationship is encoded in dependency types and dependency directions of \ac{dt}~\cite{de2008stanford}. 

Through the rules we design, every word in \ac{dt} is assigned with a node type and a parent node. We design 7 rules for \obj-\attr, 12 rules for \rel-\obj, 1 rule for \obj-\obj, 10 for \obj-\rel and 22 rules for function word processing.\footnote{The dependencies used here is based on Stanford typed dependencies \cite{de2008stanford}, a framework for annotation of grammar (parts of speech and syntactic dependencies \url{https://catalog.ldc.upenn.edu/LDC99T42})~\cite{marcus-etal-1993-building}. 
Following \cite{zhao-titov-2020-visually}, a learned parser \cite{zhangetal2020efficient} on this annotated data is used to label the dependency existence and dependency type.}

\paragraph{DT-SG Alignment} 
Based on the rewritten \ac{dt}, we perform DT-SG alignment to map the rewritten \ac{dt} to \ac{sg}. In details, 
we calculate the similarity score between the \ac{sg} node and the word’s parent, and choose top $k$ results as the alignment result. The words labeled attribute leverage parent to retrieve \obj node it attributes in \ac{sg}. Then we retrieve the \attr node in the subtree rooted by the \obj node by calculating the similarity score between the word and \attr node name. An illustration of the process of alignment from words to \ac{sg} nodes is shown in \Cref{fig:alignment}.


\subsection{Crowd-Sourcing Human Refinement}
To obtain a high-quality dataset, a human-refinement stage is adapted, providing an automatically annotated VL structure and asking the annotators to output the refined one. We utilize \ac{amt} to hire remotely native speakers to perform a crowd-sourcing survey.

\paragraph{Human Refinement}
We create a survey in \ac{amt} that allows workers to assess and refine data generated from the automatic rule-based alignment stage. We provide workers with comprehensive instructions and a set of well-defined examples to judge the quality of alignments, and modify those unsatisfied ones. During the task, we will show workers an interface with paired images and captions grouped by the image. We ask workers to check \ac{dt}, \ac{sg}, and the cross-modality alignment. Then the workers correct the inappropriate areas when necessary. The final results are combined using a majority vote.

\paragraph{Quality Control}
We adopt a set of measurements for quality control during the calibration process. Before submitting the task, the survey will first check the modifying parts by the worker to ensure that the modifying parts meet the base requirement: a dependency in \ac{dt} is aligned to a \rel node in \ac{sg}. If we find this kind of misalignment during annotation, we will prompt a message asking the workers to recheck their annotation. We publish datasets to workers one by one and request at least two workers to process the same sample to check whether there are disagreements. To ensure high-quality labeling, we restrict participated workers who have finished 500 \acp{hit} with high accuracy in the labeling history.

We do the post-processing double-check after the human refinement. We collect the flag disagreements for multiple decisions from several workers. All samples that have disagreement are double-checked manually by a third-party worker. We also flag annotations from workers whose work seems inadequate and filter out their results from the final collections.



\subsection{Dataset Analysis}
For the training dataset, we inherit MSCOCO training dataset~\cite{10.1007/978-3-319-10602-1_48}\footnote{We use the training data split following \citet{zhao2020visually}. It contains 82,783 training images, 1,000 validation images, and 1,000 test images.}. We annotate \ac{vl} structures based on the intersection of MSCOCO  \textit{dev}+\textit{test} datasets and Visual Genome~\cite{krishnavisualgenome}. We collect an annotated dataset with 850 images and 4,250 captions (each image is associated with 5 captions). Then we split the 850 images into \textit{dev} and \textit{test} datasets by 1:1. The remains in \textit{dev}+\textit{test} are merged into the training dataset. \Cref{tab:data-analysis} shows the data summary. 


 \begin{table}[ht!]
\centering
\resizebox{0.75\linewidth}{!}{%
\begin{tabular}{l|c|c|c}
\hline \toprule
                               & \textbf{Train} & \textbf{Dev} & \textbf{Test} \\ \hline
\# Images  &     83933             &       425         &     425            \\ \hline
\# Sentences  &    419665              &    2125            &   2125              \\ \hline
\# Avg. Instances in DT &        -          &     20           &   21              \\ \hline
\# Avg. Instances in SG &         -        &     135           &    134            \\ \bottomrule
\end{tabular}%
}
\caption{Data analysis of \texttt{VLParse}. \# Avg.: The average number. Instances include zero-order instances, first-order relationships, and second-order relationships.}
\vspace{-.1in}
\label{tab:data-analysis}
\end{table}

\subsection{Human Performance}
Five different workers are asked to label parse trees of 100 sentences from the test set. A different set of five workers on \ac{amt} were asked to align the visual terms and the language terms on the same sentences and their corresponding images. Then the averaged human performance is calculated as 96.15\%.

Based on these observations, our designed dataset presents language representation and cross-modality understanding clearly and keeps vision-language alignment concrete. It demonstrates the reliability of our new dataset and benchmark through manual review.

\section{Unsupervised Vision-Language Parsing}\label{sec:task}
In this section, we introduce the task of unsupervised \acf{vl} parsing, short for \texttt{VLParse}.
We formalize the task of \ac{vl} parsing followed by evaluation metrics.

\subsection{Task Formulation}
Given an input image $\mathbf{I}$ and the associating sentence with a sequence of $N$ words $\mathbf{w}=\{w_1, w_2, ..., w_N\}$, the task is to predict the joint \aclu{pt} $\mathbf{pt}$ in a unsupervised manner. Specifically, the goal is to induce VL structures from only image-caption pairs without annotations of \ac{dt}, \ac{sg} nor phrase-region correspondence annotations for training. Of note, we do use a pre-trained object detector to obtain 50 bounding boxes as candidates, while the labels of the bounding boxes are not given. For a fully unsupervised setting, the process of obtaining the bounding boxes can be replaced by an object proposal method (\eg,~\citep{Uijlings13Selective}). Compared with the weakly-supervised scene graph grounding task as in \cite{shi2021simple}, the scene graphs in \texttt{VLParse}~are unknown.
Each \obj node in language \ac{dt} will be mapped to a box region $o_i\in \mathbb{R}^4$ given $M$ candidate object proposals $\mathcal{O}=\{ o_i \}_{i=1}^M$ of the corresponding image. So are the relationships.


\subsection{Evaluation Metrics}

Due to lacking annotations of \ac{vl} structure, we indirectly assess our model by two derived tasks from each modality’s perspective, \ie, language dependency parsing and phrase grounding. 

\noindent\textbf{Directed / Undirected Dependency Accuracy (DDA/UDA)} DDA and UDA are two widely used evaluation metrics of dependency parsing. DDA denotes the proportion of tokens assigned with the correct parent node. UDA denotes the proportion of correctly predicted undirected dependency relation.

\noindent\textbf{Zero-Order Alignment Accuracy (Zero-AA)} Zero-AA assesses the alignment results on the zero-order level. A word is considered successfully grounded if two conditions are satisfied. First, the predicted bounding box of a language vertex has at least $0.5$ IoU (Intersection over Union) with the box of ground-truth SG vertex if the ground-truth is a \obj node or \attr node, or the connected two boxes both have at least $0.5$ IoU scores if the ground-truth is a \rel node. Second, although \obj node and \attr node share the same region, we ask models to distinguish them.  

\noindent\textbf{First/Second-Order Alignment Accuracy (First/Second-AA)} We are also interested in whether the first- and second-order relationships remain after alignment to another modality. That is, whether two zero-order instances (subject and predicate) in the first-order relationship remain adjacent in the aligned \ac{sg}. For the second-order relationship, we consider whether three zero-order instances (a subject, a predicate, and an object) remain adjacent. For the second-order relationships, there are multiple approach to connect the three words obj-pred-sub (\eg, obj$\rightarrow$pred$\rightarrow$sub and obj$\leftarrow$pred$\rightarrow$sub). We consider them all correct because distinguishing their adjacency to an unsupervised parser is more important to identify the semantics.



\section{Vision-Language Graph Auto-Encoder}
\label{sec:model}

\begin{figure*}[t!]
    \centering
    \includegraphics[width=\linewidth]{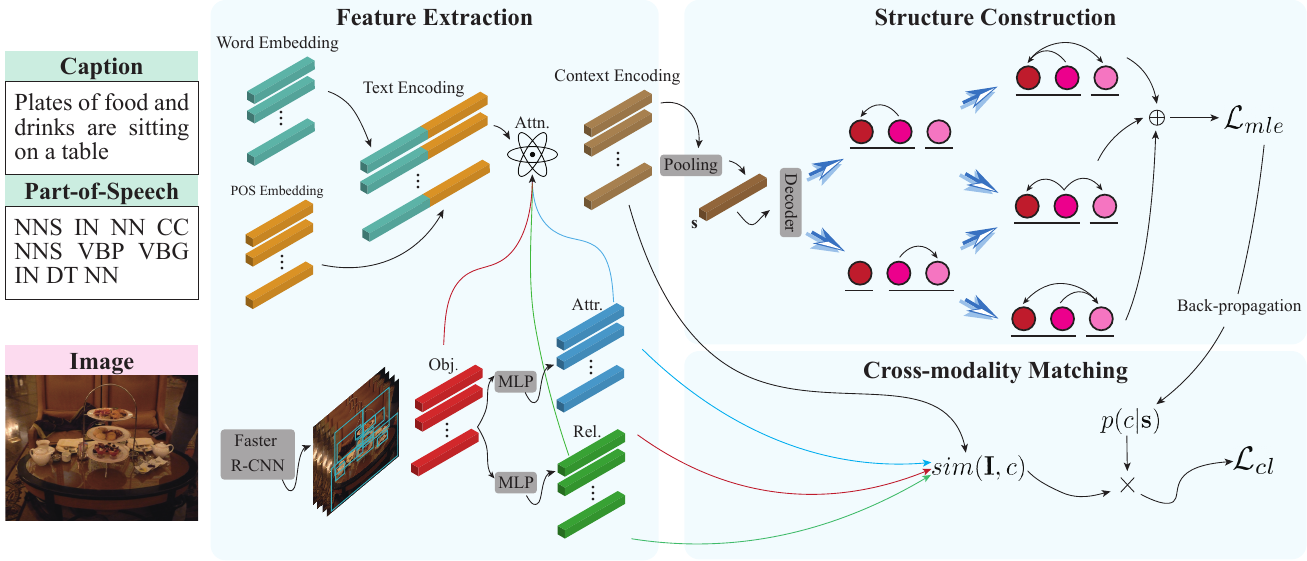}
    \caption{Diagram of VLGAE. It first extracts features from both modalities and builds representations for all instances in DT and SG. Then the encoder of the structure construction module encodes the language feature with visual cues and output a compressed representation $\mathbf{s}$. With this compressed global representation $\mathbf{s}$, the decoder incorporates the inside algorithm to construct the VL structure recursively as well as compute the posteriors. On top of the resulting posteriors generated from the structure construction module, an enhanced matching score between language context $c$ and image region $v$ is used to promote the cross-modality fine-grained correspondence.}
    \label{fig:diagram}
    \vspace{-.05in}
\end{figure*}


In this section, we introduce a novel \ac{cl} based architecture, VLGAE, to benchmark the \texttt{VLParse} task. The architecture is composed of feature extraction, structure construction and cross-modality matching modules; \Cref{fig:diagram} depicts the overall computational framework. Below we will discuss details of each module and then the learning and inference algorithms. 


\subsection{Modeling}

\paragraph{Feature Extraction} For visual features, we start by using an off-the-shelf object detector Faster R-CNN~\cite{ren2015faster} to generate a set of object proposals (RoIs) $\mathcal{O}=\{ o_i \}_{i=1}^M$ on an input image $\mathbf{I}$ and extracting corresponding features $\{v^o_i\}_{i=1}^M \in \mathbb{R}^D$ as \obj
nodes' features, where $D$ is the dimension of each RoI feature. For each \obj node $v_i^o$, an \attr node is tagged along, with its feature denoted as $v^a_i = \text{MLP}(v_i^o)$.
For two arbitrary \obj nodes $v^o_i$ and $v^o_j$, we denote the zero-order \rel node as $v_{i\rightarrow j, 0}^{\img}$. 
We also add an dummy node representing the full image and take the average of all \obj node features as its feature. For all nodes except for \obj nodes, we use randomly initialized neural networks to represent the features.

For textual features, each word $w_i$ in sentence $\mathbf{w}$ is represented as the concatenation of a pretrained word embedding $w_i$ and a randomly initialized POS tag embedding $t_i$. 
Similar to \rel nodes in \acp{sg}, the representation of dependency between two words, $w_{i\rightarrow j}$ is extracted by neural networks fed with $(w_i, w_j)$. We use Biaffine scorers~\cite{dozat2016deep} for the first-order relationship:
\begin{align*}
\vspace{-10pt}
    &w_{i}^{1st,parent}, w_{i}^{1st,child} = \text{MLP}^{1st,parent/child}(w_i) \\ 
    &w_{i\rightarrow j}^{1st} = \text{Biaffine}(w_{i}^{1st,parent}, w_{i}^{1st,child})  \\ 
    &\text{Biaffine}(w_i, w_j) = w_i^{\rm T} \mathbf{W}_{1} w_j + (w_i+w_j)^{\rm T} \mathbf{W}_2+ b,
    \vspace{-10pt}
\end{align*}
where MLP denotes the multi-layer perceptron, $\mathbf{W}_1$, $\mathbf{W}_2$ and $b$ are trainable parameters. The calculation of a second-order relationship's score follows a similar way. 

\paragraph{Structure Construction}
Inspired by neural \ac{dt} construction algorithms~\cite{han2019enhancing}, we use an encoder-decoder framework that employs the dynamic programming algorithm (namely, inside algorithm) and calculate the posteriors of instances $p(\mathbf{pt}|\mathbf{w}, \mathbf{I})$ recursively retrieved during the structure construction.

\noindent\emph{\underline{Encoder}}\quad{}
The encoder is to produce a joint representation of an input image $\mathbf{I}$ and its corresponding caption $\mathbf{w}$. Specifically, we obtain contextual encoding $c\in \mathcal{C}$ by fusing the text features with the visual information via attention mechanisms, where $\mathcal{C}$ denotes the space for the attended language context. 
For each token in captions $\{w_i\}$ and \ac{sg} representations $\mathcal{V}=\{v_i, v_{i\rightarrow j}\}$, we calculate attention scores between them and then obtain weighted summation over all terms, \ie, $c_i = \sum {\rm Attn}(w_i, v_i) w_i$.
Finally, we use an average-pooling layer to summarize all information into a continuous context vector $\mathbf{s}$, which represents the global information of the vision-language context.


\noindent\emph{\underline{Decoder}}\quad{} 
The decoder generates the tag sequence $\mathbf{t}$ and parse tree $\mathbf{pt}$ conditioned on the joint representation $\mathbf{s}$ \wrt the joint probability $p(\mathbf{t}, \mathbf{pt}|\mathbf{s})$.
To consider the exponential scale of possible parse trees, we use dynamic programming to consider all possible dependencies over the sentence. 
Refer to \Cref{sec:learning} for the learning process. 

\paragraph{Cross-modality Matching}
We employ cross-modality matching to align vision and language features in different levels.

\noindent\emph{\underline{Matching Score}}\quad{} 
We define $sim(\cdot,\cdot)$ as the cross-modality matching function. Following \citet{wang2020maf}, we first compute the similarity score between each $c \in \mathcal{C}$ and each $v \in \mathcal{V}$:
\begin{align}\label{eq:sim_v_h}
	sim(v, c) =  \langle v , c  \rangle,
\end{align}
where $\langle \cdot, \cdot \rangle$ is an inter-product function. 
Heuristically, we can define the similarity score between instance $c$ and the entire image $\mathbf{I}$ as
\begin{align}
	sim(\mathbf{I}, c) = \max_{v \in \mathcal{V}} sim(v, c),
\end{align}

\noindent\emph{\underline{Matching Score Enhanced by Posterior}}\quad{} To leverage the contextual information, we use a posterior $p(c|\mathbf{s})$ computed from the decoder to reflect how likely $c$ exists given the joint representation $\mathbf{s}$. 
Then we fuse the matching scores with the posteriors to provide an enhanced simliarity function,  $sim^+(\mathbf{I}, c) = sim(\mathbf{I}, c) \times p(c|\mathbf{s})$.

\subsection{Learning}
\label{sec:learning}
\paragraph{\acf{mle}}

With the compressed representation $\mathbf{s}_i$ for image-sentence pair $(\mathbf{I}_i, \mathbf{w}_i)$, VLGAE generates the tag sequence $\mathbf{t}_i$ and the parse tree $\mathbf{pt}$. 
The learning objective is to maximize the conditional log-likelihood of $K$ training sentences:
\begin{equation}
\begin{aligned}
\mathcal{L}_{mle} &= - \frac{1}{K}\sum\limits_{i=1}\limits^{K}\log p_{\Theta}(\mathbf{t}_i | \mathbf{w}_{i}) \\
&= -\frac{1}{K}\sum\limits_{i=1}\limits^{K}\log \sum_{\mathbf{pt} \in PT(\mathbf{s}_i)} p_{\Theta}( \mathbf{t}_{i}, \mathbf{pt} | \mathbf{w}_{i}) 
\end{aligned}
\label{eqn:mle}
\end{equation}   
where $\Theta$ parameterizes the encoder-decoder neural network and $PT(\mathbf{s}_i)$ denotes the set of all possible parse trees.
Given some $\Theta$, \Cref{eqn:mle} can be computed using the inside algorithm, an $\mathcal{O}(n^3)$ dynamic programming procedure. Therefore, we perform structure construction and parameter learning via an \ac{em} process. Specifically, the E-step is to compute possible structures given current $\Theta$ and the M-step is to optimize $\Theta$ by gradient descent \wrt \Cref{eqn:mle}. Of note, the posterior $p(c|\mathbf{s})$ used for matching score can be computed in the back-propagation process~\cite{eisner-2016-inside}. 

\paragraph{Contrastive Loss}
Due to the lack of fine-grained annotations in an unsupervised setting, the objective referring to the alignment is employed in a contrastive loss. The contrastive learning strategy is based on maximizing the matching score between paired fine-grained instances. For each $c$, the sentence's corresponding image is a positive example and all the other images in the current batch are negative examples. Of note, compared with coarse image-sentence pairs, our design of fine-grained vision-language alignments yield stronger negative pairs for contrastive training. 
Formally, given a vision-language pair $(\mathbf{w}, \mathbf{I})$ within a batch, the contrastive loss can be defined as,
\begin{align}
\mathcal{L}_{cl}(\mathbf{w}, \mathbf{I}) &= \mathbb{E}_{p(\mathbf{pt}|\mathbf{w})} \sum_{c\in \mathbf{pt}} \ell(\mathbf{I}, c), \\
 \ell(\mathbf{I}, c) =& -\log \frac{\exp [sim^{+}(\mathbf{I}, c)]}{\sum_{\hat{\mathbf{I}}\in batch} \exp [sim^{+}(\hat{\mathbf{I}}, c)]},
\label{eq:cl}
 \end{align}
 where $\mathbf{pt}$ is a valid parse tree, $\hat{\textbf{I}}$ are negative examples in a batch. $\ell(\mathbf{I}, c)$ shows an possibly aligned pair that ranks higher than other unaligned ones in a batch.

Finally, the total loss is defined as
\begin{equation}
\begin{split}
	\mathcal{L}_{tot} &= (1-\lambda)\cdot\mathcal{L}_{mle} + \lambda \cdot \mathcal{L}_{cl},
\end{split}
\end{equation}
where $\lambda$ is pre-defined to balance different scalars between two losses. 

\subsection{Inference}
\label{sec:inference}
Given a trained model with trained parameters $\boldsymbol{\Theta}$, the model can predict the VL structure and further the parse tree of the sentence and its visual grounding on the \ac{sg}. The parse tree can be parsed by searching for $\mathbf{pt}^*$ with the highest conditional probability among all valid parse trees $PT(\mathbf{s})$ using the dynamic programming  \cite{klein2004corpus}: 
\begin{equation}
\mathbf{pt}^{*} = \argmax_{\mathbf{pt} \in  PT(\mathbf{s})}~p(\mathbf{pt} | \mathbf{s}; \boldsymbol{\Theta})
\end{equation}

For each $c \in \mathcal{C}$, we can predict its corresponding image region $o_{m^{c}}$ using enhanced similarity score as in \Cref{eq:sim_v_h}:
\vspace{-2pt}
\begin{align}
	v^{*} =  \argmax_v~sim^{+}(v, c) \label{eq:word-ground-enhance}
\vspace*{-2pt}
\end{align}
It is worth noting that, when the ground truth dependency tree is known for sentence, we can directly retrieve corresponding scene graph \wrt \Cref{eq:word-ground-enhance}.

\section{Experiments}
\label{sec:exp}

\subsection{Setup}
The candidate bounding boxes are given in the following setting.
For an input image, we use an external object detector, Faster R-CNN as MAF~\cite{wang2020maf}, to generate top-50 object proposals.
For each proposal, we use RoI-Align~\cite{he2017mask} and global average pooling to compute the object feature \cite{wang2020maf}.
Since we do not have the ground-truth structure of the captions, we follow \cite{shi2019visually} and \cite{zhao-titov-2020-visually} to use predictions as ground truth produced by an external parser. We report the average score of three runs with different random seeds. 



\subsection{Evaluation on Language Structure Induction}

\begin{table}[t!]
\centering
\resizebox{0.25\textwidth}{!}{%
\begin{tabular}{rcc}
\bottomrule[1pt]
\multicolumn{1}{r|}{}                               & \textbf{UDA} & \textbf{DDA}              \\ \hline
\multicolumn{3}{c}{\textit{Language Only}}                                                               \\ \hline
\multicolumn{1}{r|}{Left branch}                    &   53.61         &  30.75                            \\
\multicolumn{1}{r|}{Right branch}                   &  53.19       & 23.01                             \\
\multicolumn{1}{r|}{Random}                         &  32.44     & 19.29                                 \\
\multicolumn{1}{r|}{DMV~\cite{klein2004corpus}}                         &  58.06    &  41.36                           \\
\multicolumn{1}{r|}{D-NDMV~\cite{han2019enhancing}}                          &     70.77           &   65.88                           \\\hline
\multicolumn{3}{c}{\textit{Vision-Language (VL)}}                                                                      \\ \hline
\multicolumn{1}{r|}{\textbf{VLGAE}} &    \textbf{71.43}    & \textbf{67.57}  \\ \bottomrule[1pt]
\end{tabular}%
}
\caption{Dependency structure induction results on the test split.} 
\vspace{-.05in}
\label{tab:langauge_parsing}
\end{table}
We compare VLGAE with prior language-only baselines on language structure induction using UDA and DDA metrics in \Cref{tab:langauge_parsing}. We can obverse a performance boosting after incorporating visual cues. In particular, VLGAE outperforms D-NDMV by 1.69\%  score on DDA and 0.66\% score on UDA.

\subsection{Evaluation on Visual Phrase Grounding}
In addition to language structure induction, we evaluate our approach on the weakly-supervised visual phrase grounding task.\footnote{We apply the learning strategy of MAF on weakly-supervised visually grounding of a DT instead of given noun phrases in the training stage. } Experimental results in \Cref{tab:grounding} show that VLGAE outperforms the previous mutlimodal baseline MAF~\cite{wang2020maf} by $1.0\%$.
Moreover, a significant improvement is observed, especially for high-order relations, indicating the effectiveness of our multi-order alignments. We also report performance if ground truth bounding boxes (and relationships) are used as a reference instead of proposals (and dense connections); see VLGAE$^\dagger$ in \Cref{tab:grounding}.

\subsection{Ablation Analysis on Arc Length}\label{sec:analysis_length}
\begin{figure}[th!]
    \centering
    \includegraphics[width=.8\linewidth]{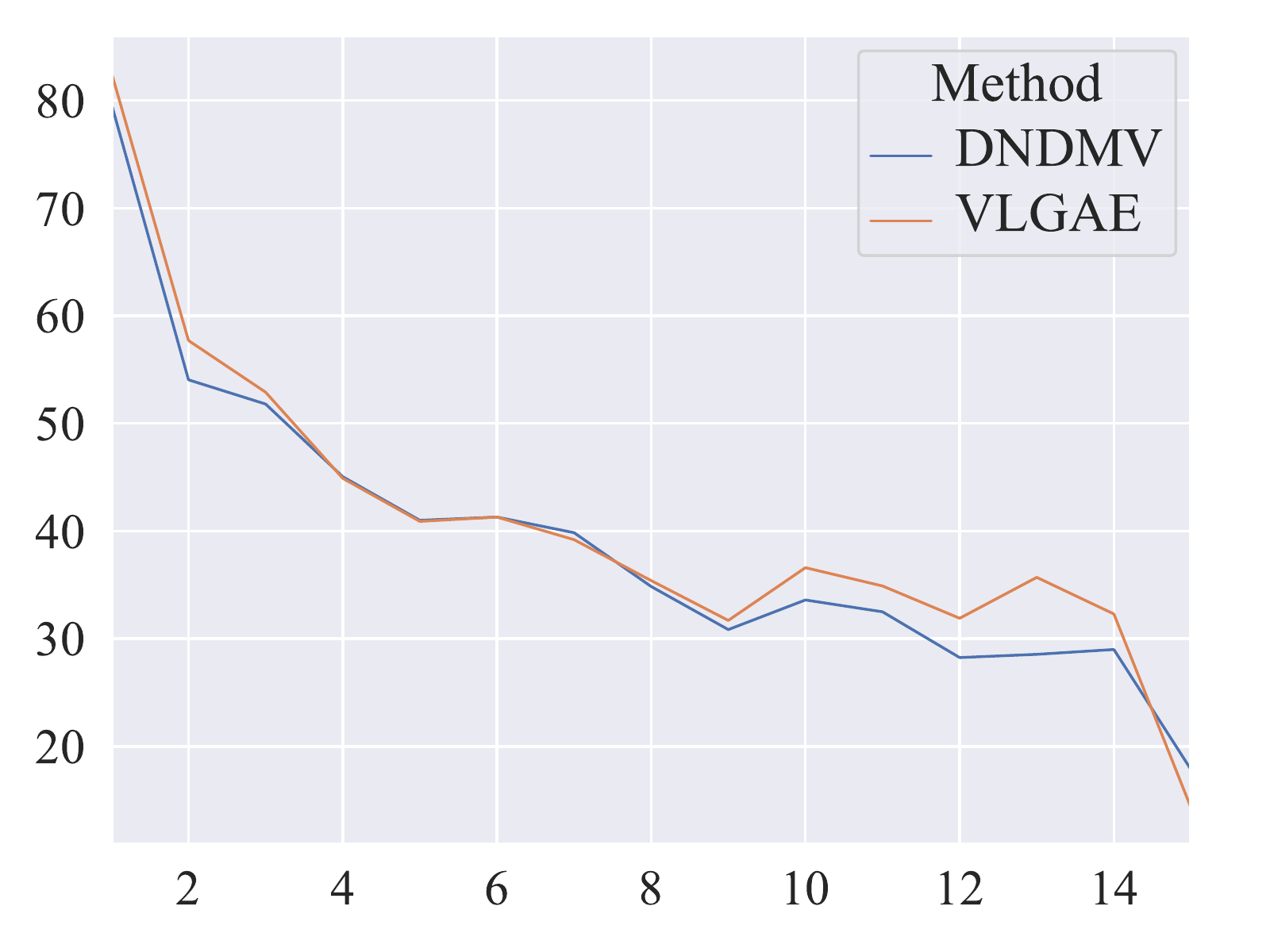}
    \caption{DDA of different arc length on the Dev dataset.}
        \vspace{-.05in}
    \label{fig:analysis_on_length}
\end{figure}

We further investigate the recall rate for different lengths of arcs $len(w_{i \rightarrow j})$ in \Cref{fig:analysis_on_length}. The experiments are on the Dev split. VLGAE enhanced by visual cues has been proven to boost DDA/UDA than its non-visual version (D-NDMV) in \Cref{tab:langauge_parsing}. Moreover, this boost is observed not only on short arcs but also longer arcs. This phenomenon is contrary to VC-PCFG \cite{zhao-titov-2020-visually}, showing that dependency structures in VLGAE can be beneficial for all the arcs regardless of the arc length, compared with constituent structures.

\begin{table}[t!]
\centering
\resizebox{0.9\linewidth}{!}{%
\begin{tabular}{cccccccc}
\toprule[1pt]
  &\textit{All}                   & \textit{Obj.} & \textit{Attr.} & \textit{Rel.} & \textit{First}                      & \textit{Second}                    \\ \hline
\textbf{Random} & 12.2 &   15.9          &    9.4               &  0.0                                               &   0.0                             &    0.0                             \\
\textbf{MAF*}   & 27.7 &    38.5           &     20.7               &                0.1                                   &       0.0                        &  0.0                       \\
\textbf{VLGAE}   & \textbf{28.7}    &   \textbf{36.1}      &   \textbf{21.0}             &   \textbf{10.2}                                         &       \textbf{3.4}                      & \textbf{0.2}                        \\ \hline
VLGAE$^\dagger$ & 42.3 & 67.2 & 41.8 & 15.9 & - & - \\
\bottomrule[1pt]
\end{tabular}%
}
\caption{Visual grounding results on the test split. $*$ refers to re-implemented results. $\dagger$ refers to experiments using gold scene graphs. \textit{All}: Zero-AA on all zero-order instances. \textit{Obj.}: Zero-AA on objective nodes. \textit{Attr.}:Zero-AA on attribute nodes. \textit{Rel.}: Zero-AA on relationship nodes. \textit{First}: First-AA. \textit{Second}: Second-AA.}
\label{tab:grounding}
\vspace{-.05in}
\end{table}


\section{Conclusion}\label{sec:conclusion}

In this work, we introduce a new task \texttt{VLParse} that aims to construct a joint \ac{vl} structure that leverages both visual scene graphs and language dependency trees in an unsupervised manner. Meanwhile, we deliver a semi-automatic strategy for creating a benchmark for the proposed task. Lastly, we devise a baseline framework VLGAE based on contrastive learning, aiming to construct such structure and build \ac{vl} alignment simultaneously. Evaluations on structure induction and visually phrase grounding show that VLGAE enhanced by visual cues can boost performance than its non-visual version. Despite of the compelling boosted results, the performance on both tasks are far from satisfactory. Nevertheless, this work sheds light on explainable multimodal understanding and calls for future research in this direction.


{\small
\bibliographystyle{ieee_fullname_natbib}
\bibliography{acl_anthology,egbib}
}


\end{document}